%% file: musst.tex
\def\year{2021}\relax
\documentclass[letterpaper]{article} % DO NOT CHANGE THIS
\usepackage{aaai21}  % DO NOT CHANGE THIS
\usepackage{times}  % DO NOT CHANGE THIS
\usepackage{helvet} % DO NOT CHANGE THIS
\usepackage{courier}  % DO NOT CHANGE THIS
\usepackage[hyphens]{url}  % DO NOT CHANGE THIS
\usepackage{graphicx} % DO NOT CHANGE THIS
\usepackage{natbib}  % DO NOT CHANGE THIS AND DO NOT ADD ANY OPTIONS TO IT
\usepackage{caption} % DO NOT CHANGE THIS AND DO NOT ADD ANY OPTIONS TO IT
\usepackage{algorithm}
\usepackage{algpseudocode}
\usepackage{booktabs}
\usepackage{multirow}
\usepackage{multicol}
\usepackage{arydshln}
\usepackage{pbox}
\usepackage{soul}

\input{math_commands.tex}

\newcommand{\spade}{\textsuperscript{$\spadesuit$}} 
\newcommand{\td}{\textsuperscript{\textdagger}}

\title{Multi-span Style Extraction for Generative Reading Comprehension}
\author{

    Junjie Yang\textsuperscript{\rm 1,3,4},
	Zhuosheng Zhang\textsuperscript{\rm 2,3,4},
	Hai Zhao\textsuperscript{\rm 2,3,4\thanks{Corresponding author.  This paper was partially supported by National Key Research and Development Program of China (No. 2017YFB0304100), Key Projects of National Natural Science Foundation of China (U1836222 and 61733011), Huawei-SJTU long term AI project, Cutting-edge Machine reading comprehension and language model.}} 
	\\
}
\affiliations{
    \textsuperscript{\rm 1}SJTU-ParisTech Elite Institute of Technology, Shanghai Jiao Tong University\\
	\textsuperscript{\rm 2}Department of Computer Science and Engineering, Shanghai Jiao Tong University\\
	\textsuperscript{\rm 3}Key Laboratory of Shanghai Education Commission for Intelligent Interaction\\
	and Cognitive Engineering, Shanghai Jiao Tong University, Shanghai, China\\
	\textsuperscript{\rm 4}MoE Key Lab of Artificial Intelligence, AI Institute, Shanghai Jiao Tong University, Shanghai, China \\
	{\tt jj-yang@sjtu.edu.cn, zhangzs@sjtu.edu.cn, zhaohai@cs.sjtu.edu.cn} \\}
\begin{document}

\maketitle

\begin{abstract}
	Generative machine reading comprehension (MRC) requires a model to generate well-formed answers. For this type of MRC, answer generation method is crucial to the model performance. However, generative models, which are supposed to be the right model for the task, in generally perform poorly. At the same time, single-span extraction models have been proven effective for extractive MRC, where the answer is constrained to a single span in the passage. Nevertheless, they generally suffer from generating incomplete answers or introducing redundant words when applied to the generative MRC. Thus, we extend the single-span extraction method to multi-span, proposing a new framework which enables generative MRC to be smoothly solved as multi-span extraction. Thorough experiments demonstrate that this novel approach can alleviate the dilemma between generative models and single-span models and produce answers with better-formed syntax and semantics.
\end{abstract}

\section{Introduction}
Machine Reading Comprehension (MRC) is considered as a nontrivial challenge in natural language understanding. Recently, we have seen continuous success in this area, partially benefiting from the release of massive and well-annotated datasets from both academic \cite{rajpurkar_know_2018, reddy_coqa_2019} and industry \cite{bajaj_ms_2018, he_dureader_2018} communities.

The widely used span-extraction models \cite{seo_bidirectional_2017, ohsugi_simple_2019, lan_albert_2020}, formulate the MRC task as a process of predicting the start and end position of the span inside the given passage. They have been proven effective on the tasks which constrain the answer to be an exact span in the passage \cite{rajpurkar_know_2018}. However, for generative MRC tasks whose answers are highly abstractive, the single-span extraction based methods can easily suffer from incomplete answers or redundant words problem. Thus, there still exists a large gap between the performance of single-span extraction baselines and human performance.

\begin{figure}[t]
	\centering
	\includegraphics[width=0.9\linewidth]{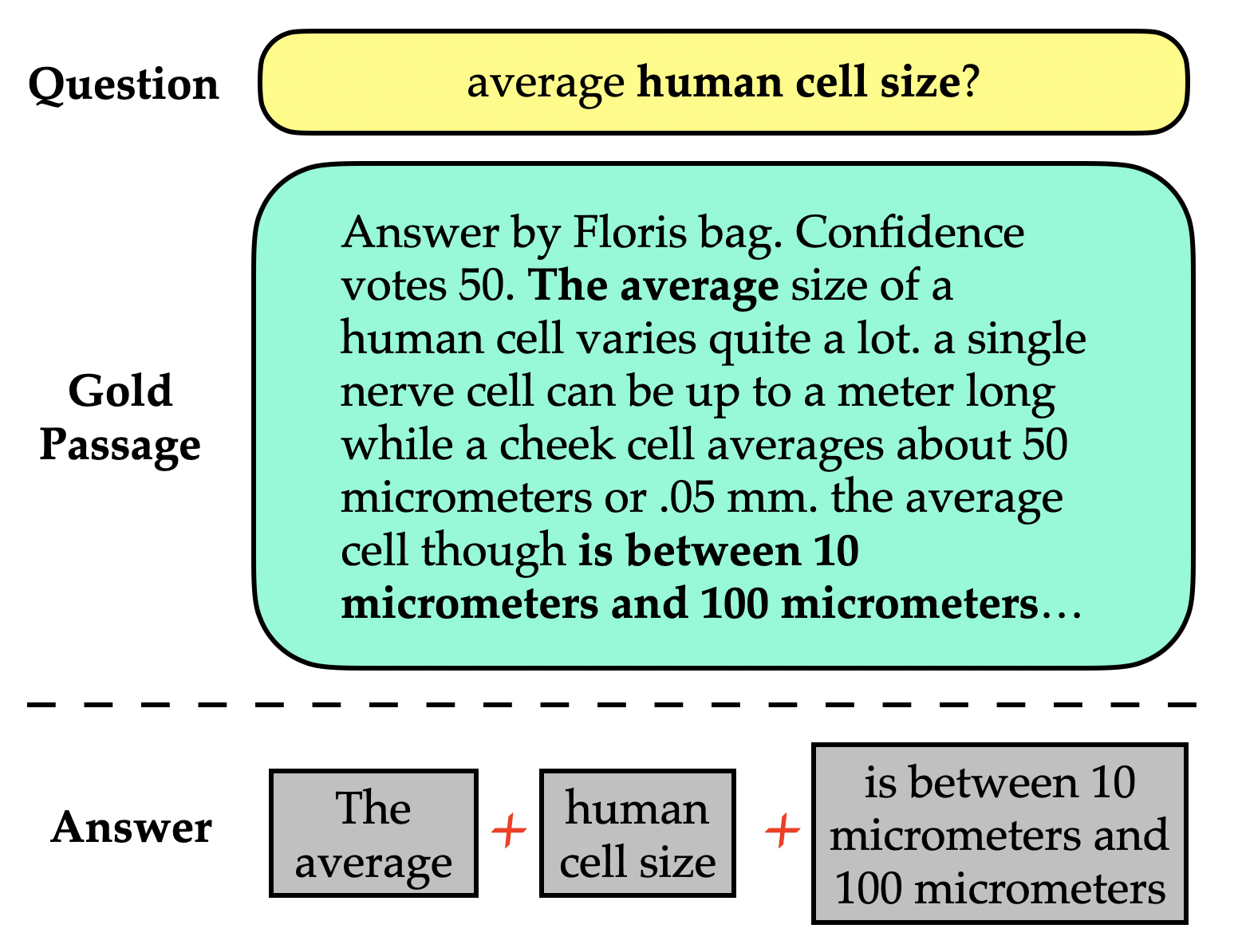}
	\caption{Example of how a well-formed answer is generated by the multi-span style extraction.}
	\label{fig:example_multi}
\end{figure}

In the meantime, we have observed that utilizing multiple spans appearing in the question and passage to compose the well-formed answer could be a promising method to alleviate these drawbacks. Figure \ref{fig:example_multi} shows how the mechanism of multi-span style extraction works for an example from the MS MARCO task \cite{bajaj_ms_2018}, where the well-formed answer cannot simply be extracted as a single span from the input text.

\begin{figure*}[ht]
	\includegraphics[width=\linewidth]{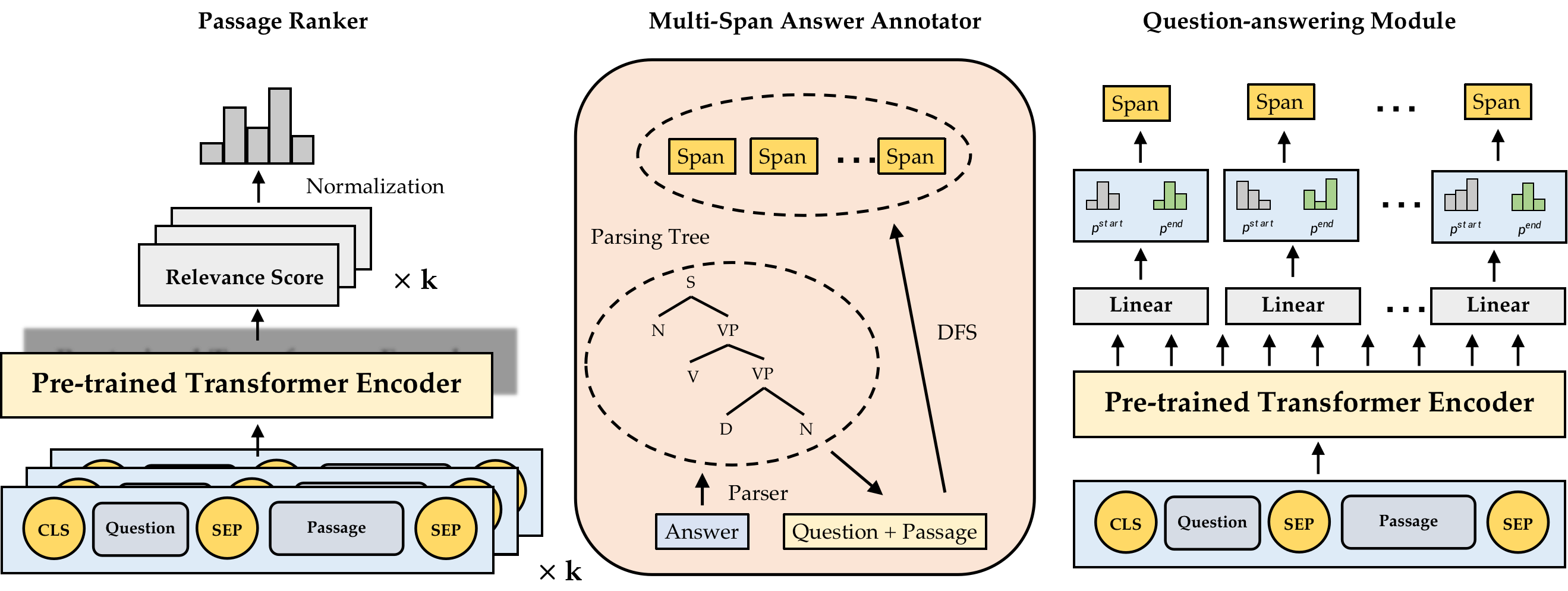}
	\caption{Our framework MUSST}
	\label{fig:framework}
\end{figure*}

Therefore, in this work, we propose a novel answer generation approach that takes advantage of the effectiveness of span extraction and the concise spirit of multi-span style to synthesize the free-formed answer, together with a framework as a whole for the multi-passage generative MRC. We call our framework MUSST for \textbf{MU}lti-\textbf{S}pan \textbf{ST}yle extraction. Our framework is also empowered by well pre-trained language model as encoder component of our model. It provides deep understanding of both the input passage and question, and models the information interaction between them. We conduct a series of experiments and the corresponding ablations on the MS MARCO v2.1 dataset.

Our main contributions in this paper can be summarized as follows\footnote{The code is publicly available at: \url{https://github.com/chunchiehy/musst}}:
\begin{itemize}
	\item We propose a novel multi-span answer annotator to transform the initial well-formed answer into a series of spans that distribute in the question and passage.
	\item We generalize the single-span extraction based method to the multi-span style by introducing a lightweight but powerful answer generator, which supports the extraction of various number answer spans during prediction.
	\item To make better usage of the large dataset for the passage ranking task, we propose dynamic sampling during the training of the ranker that selects the passage most likely to entail the answer.
\end{itemize}

\section{MUSST}
In this section, we present our proposed framework, MUSST, for multi-passage generative MRC task. Figure \ref{fig:framework} depicts the general architecture of our framework, which consists of a passage ranker, a multi-span answer annotator, and a question-answering module.

\subsection{Passage ranker}
\subsubsection{Problem formulation}
Given a question $Q$ and a set of $k$ candidate passages $\mathbf{P}= \{P_1, P_2, ..., P_k\}$, the passage ranker is responsible for ranking the passages based on their relevance to the question. In other words, the model is requested to output conditional probability distribution $P(y|Q, \mathbf{P}; \vtheta)$, where $\vtheta$ is the model parameters and $P(y = i|Q, \mathbf{P}; \vtheta)$ denotes the probability that passage $P_i$ can be used to answer question $Q$.

\subsubsection{Encoder}
\label{encoding_layer}
For each input question and passage pair $(Q, P_i)$, we represent it as a single packed sequence of length $n$ of the form ``\texttt{[CLS]}$Q$\texttt{[SEP]}$P_i$\texttt{[SEP]}''. We pass the whole sequence into a contextualized encoder, thereby to produce its contextualized representation $\mE \in \mathbb{R}^{n\times h}$ where $h$ denotes the hidden size of the Transformer blocks. Following the fine-tuning strategy of \citeauthor{devlin_bert:_2019}~(\citeyear{devlin_bert:_2019}) for the classification task, we consider the final hidden vector $\vc \in \mathbb{R}^{h}$ corresponding to the first input token (\texttt{[CLS]}) as the input's aggregate representation. Our encoder also models the interaction between the question and the passage.

\subsubsection{Ranker}
The ranker is responsible for ranking the passages based on its relevance to the question. Given the output of the encoding layer $\vc$, we pass it through a fully connected multi-layer perceptron which consists of two linear transformations with a Tanh activation in between:
\begin{gather*}
	\vs = \text{softmax}(\mW_2\tanh(\mW_1\vc + \vb_1) + \vb_2) \in \mathbb{R}^2\\
	u_i = s_0 \text{ and } r_i = s_1
\end{gather*}
where $\mW_1 \in \mathbb{R}^{h\times h}$, $\mW_2 \in \mathbb{R}^{2\times h}$, $\vb_1 \in \mathbb{R}^{h}$ and $\vb_2 \in \mathbb{R}^2$
are trainable parameters. Here, $r_i $ and $u_i$ are respectively the relevance and unrelevance score for the pair $(Q, P_i)$. The relevance scores are consequently normalized across all the candidates passages of the same question:
$$\hat{r}_i = \frac{\exp{(r_i)}}{\sum_{j=0}^{k}\exp{(r_j)}}$$
Here, $\hat{r}_i$ indicates the probability that passage $P_i$ entails the answer $Q$.

\subsubsection{Training}
We define the question-passage pair where the passage entails the question as a positive training sample. The positive passage is noted as $P^+$. During the training phase, we adopt a negative sampling with one negative sample. Specifically, for each positive instance $(Q, P^+)$, we randomly sample a negative passage $P^{-}$ from the \textit{unselected passages} of the same question.
The model is trained by minimizing the following cost function:
$$J(\vtheta) = -\frac{1}{T} \sum_{t=1}^{T} \log(r(Q_t, P_t^+)) + \log(u(Q_t, P_t^-))$$
where $T$ is the number of questions in the training set, $r(Q_t, P_t^+)$ denotes the relevance score of $(Q_t, P_t^+)$ and $u(Q_t, P_t^-)$ denotes the unrelevance score of $(Q_t, P_t^-)$.

Moreover, motivated by \citeauthor{liu_roberta:_2019}~(\citeyear{liu_roberta:_2019}), we resample the negative training instances at the beginning of each training epoch, to avoid using the same training pattern for the question during each training epoch. We name it \textbf{dynamic sampling}.

\subsection{Syntactic multi-span answer annotator}
\begin{algorithm}[H]
	\caption{Syntactic Multi-span Answer Annotation}
	\label{alg:annotation}
	\textbf{Input}: Question $Q = \{q_1, q_2,\dots, q_m\}$ , passage $P = \{p_1, p_2,\dots, p_n\}$ and gold answer $A= \{a_1, a_2,\dots, a_k\}$\\
	\textbf{Parameter}: Edit distance threshold $d_{max}$\\
	\textbf{Output}: A list of start and end position of answer spans in the question and passage
	\begin{algorithmic}[1] %[1] enables line numbers
		\State Let $M$ be an empty list
		\State Pack question $Q$ and passage $P$ into a single sequence $C$ in a certain way.
		\State Get the syntactic parsing tree $\gT$ of answer $A$ by a constituency parser.
		\State Let $\sS$ be the stack of subtrees to be traversed.
		\State Initialize $\sS$ with the root $\gR$ of the tree $\gT$
		\While{$\sS$ is not empty}
		\State let $\gV = $ \Call{Pop}{$\sS$}
		\State Get a list of all the leaves of subtree $\gV$: $L = \{l_1, l_2,\cdots, l_n\}$
		\If{$L$ is a sublist of $C$}
		\State Get the start index $s$ and end index $e$ of $L$ in $C$ by Knuth-Morris-Pratt pattern searching algorithm
		\State Add ($s$, $e$) into the span position list $M$
		\Else
		\For{childtree $\gU$ in $\gV$ (From right to left)}
		\State \Call{Push}{$\sS$, $\gU$}
		\EndFor
		\EndIf
		\EndWhile
		\State Reconstruct answer $A'$ from span position list $M$
		\State Let $d = $ \Call{EditDistance}{$A$, $A'$}
		\If{$ d > d_{max}$}
		\State Empty the list $M$
		\EndIf
		\State $M^*$ = \Call{Pruning}{$M$}
		\State \textbf{return} $M^*$
	\end{algorithmic}
\end{algorithm}

In this section, we introduce our syntactic multi-span answer annotator. Before the training of our question-answering module, we need to extract non-overlapped spans from the question and passage based on the original answer from the training dataset. Our annotator is responsible for transforming the original answer phrase into multiple spans that distribute in the question and passage with subject to syntactic constraints. The attempt to extract the answer spans syntactically is motivated by our first intuition that the human editors compose the original answer in an analogous way.

As shown in the middle of Figure \ref{fig:framework}, we transform the answer phrase into a parsing tree and traverse the parsing tree in a DFS (Depth-first search) way. At each visit of the subtree, we check if the span represented by the subtree appears in the question or passage text. We obtain a span list after traversing the whole parsing tree. However, in some cases, the original answer still cannot be perfectly composed by the words from the input text even in a multi-span style. We get rid of these \textit{bad} samples by comparing their edit distances with a threshold value which is set by the model beforehand.

An important final step is to prune the answer span list. The \textbf{pruning} procedure sticks to the following principle: if two spans adjoint in the list are contiguous in the original text, we joint them together. Pruning reduces heavily the number of spans needed to recover to the original answer phrase. The more comprehensive detail of our annotator is described in Algorithm \ref{alg:annotation}.

\subsection{Question-answering module}
\subsubsection{Problem formulation}
Given a question $Q$ and a passage $P$, the question-answering module is requested to answer the question based on the information provided by the passage. In other words, the model outputs the conditional probability distribution $P(y|Q, P)$, where $P(y = A|Q, P)$ denotes the probability that $A$ is the answer.

\subsubsection{Question-passage reader}
The architecture of the reader is analogous to the encoder module of the ranker in section \ref{encoding_layer}, where we take a pre-trained language model as encoder. But instead of getting only the aggregate representation, we pass the whole output of the last layer to predict the answer spans as the follows:
$$\mM = \text{Encoder}(Q, P) \in \mathbb{R}^{h \times n} $$
where $n$ is the length of the input token sequence, and $h$ is the hidden size of the encoder.

\subsubsection{Multi-span style answer generator}
Our answer generator is responsible for composing the answer in a multi-span style extraction. Let $n$ be the number of span to be extracted.

For each single span prediction, we treat it as the single span extraction MRC task. Following \citeauthor{lan_albert_2020}~(\citeyear{lan_albert_2020}), we adopt a linear layer to predict start and end positions of the span in the input sequence. It is worth noticing that our model is also enabled to predict the answer span from the question. The probability distribution of $i$-th span's start position over the input tokens is obtained by:
$$\hat{\vp}^{j,\text{start}} = \text{softmax}(\mW_j^{s}\mM + b_j^{s})$$
where $\mW_j^{s} \in \mathbb{R}^{1 \times h}$ and $b_j^{s} \in \mathbb{R}$ are trainable parameters and $\hat{\vp}_{k}^{j,\text{start}}$ denote the probability of token $k$ being the start of the answer span $j$. The end position distribution of the answer span $j$ is obtained by using the analogous formula:
$$\hat{\vp}^{j,\text{end}} = \text{softmax}(\mW_j^{e}\mM +b_j^{e})$$

\subsubsection{Training and inference}
During training, we add a special virtual span, with start and end position values equaling the length of the input sequence, at the end of the annotated answer span list. This approach enables our model to generate a various number of answer spans during prediction with the virtual span serving as a stop symbol.
The cost function is defined as follows:
$$J(\vtheta) = -\frac{1}{T}\sum_{t=1}^{T}\sum_{j=1}^{m_t}\log(\hat{\vp}_{y_t^{j,\text{start}}}^{j,\text{start}}) + \log(\hat{\vp}_{y_t^{j,\text{end}}}^{j,\text{end}})$$
where $T$ is the number of training samples, $m_t$ is the number of answer span for sample $t$, $y_t^{j,\text{start}}$ and $y_t^{j,\text{end}}$ are the true start and end position of the $t$-th sample's $j$-th span.

During inference, at each time step $j$, we choose the answer span $(k, l)$ where $k < l$ with the maximum value of $\hat{\vp}_k^{j,\text{start}}\hat{\vp}_l^{j,\text{end}}$. The decoding procedure terminates when the stop span is predicted. Sometimes, the model tends to generate repeatedly the same spans. In order to alleviate the repeating problem, at each prediction time step $j$, we mask out the predicted span positions of previous time steps ($< j$) during the calculation of probability distribution of new start and end positions. Since the masking depends on the previously predicted spans, we name it as \textbf{conditional masking}. The extracted spans are later joined together to form a final answer phrase.

\section{Experiments}
\subsection{Dataset}
We evaluate our framework on the MS MARCO v2.1 \footnote{The datasets can be obtained from the official site (\url{https://microsoft.github.io/msmarco/})} \cite{bajaj_ms_2018}, which is a large scale open-domain generative task. MS MARCO v2.1 provides two MRC tasks: Question Answering (\textbf{QA}) and Natural Langauge Generation (\textbf{NLG}). The statistics of the corresponding datasets' size are presented in Table \ref{tab:data_stats}. Both datasets consist of sampled questions from Bing's search logs, and each question is accompanied by an average of ten passages that may contain the answers. $\mathcal{QA}$ and $\mathcal{NLG}$ are subsets of $\mathcal{ALL}$, which also contains the unanswerable questions.

Distinguished with the QA task, the NLG task requires the model to provide the well-formed answer, which could be read and understood by a natural speaker without any additional context. Therefore NLG-style answers are more abstract than the QA-style answers. Table \ref{tab:data_stats} shows also the percentage of examples where the answer can be extracted as a single span in the gold passage. Unsurprisingly, the answers from the $\mathcal{QA}$ set are much more likely to match a span in the passage than the ones in the $\mathcal{NLG}$ set.
Moreover, \citeauthor{nishida_multi-style_2019}~(\citeyear{nishida_multi-style_2019}) states that the QA task prefers the answer to be more concise than in the NLG task, averaging 13.1 words, while the latter one averages 16.6 words. Therefore, the $\mathcal{NLG}$ set is more suitable to evaluate model performance on generative MRC.

BLEU-1 \cite{papineni_bleu_2002} and ROUGE-L \cite{lin_rouge_2004} are adopted as the official evaluation \footnote{The official evaluation scripts can be found in \url{https://github.com/microsoft/MSMARCO-Question-Answering/tree/master/Evaluation}} metrics to evaluate model performance, while the official leaderboard chooses ROUGE-L as the main metric. In the meantime, we use Mean Average Precision (MAP) and Mean Reciprocal Rank (MRR) for our ranker.

\begin{table}[!htbp]
	\centering
	\resizebox{0.48\textwidth}{!}{
		\begin{tabular}{lccc}
			\toprule
			Dataset         & Train             & Dev              & Test    \\
			\midrule
			$\mathcal{ALL}$ & 808,731           & 101,093          & 101,092 \\
			\midrule
			$\mathcal{QA}$  & 503,370 (63.39\%) & 55,636 (45.40\%) & --      \\
			\midrule
			$\mathcal{NLG}$ & 153,725 (12.57\%) & 12,467 (24.99\%) & --      \\
			\bottomrule
		\end{tabular}}
	\caption{Statistics of MS MARCO v2.1 dataset. The numbers in parenthesis indicate the percentage of examples whose answer is single span in gold passage.}
	\label{tab:data_stats}
\end{table}

\subsection{Baseline models}
We compare our MUSST with the following baseline models: single-span extraction and seq2seq. For the single-span extraction baseline, we employ the model for the SQuAD dataset from ALBERT \cite{lan_albert_2020}. The model is trained only with samples where the answer is a single span in the passage. In the meantime, We adopt the Transformer model from \citeauthor{vaswani_attention_2017}~(\citeyear{vaswani_attention_2017}) as our seq2seq baseline. For a fair comparison, the baseline models share the same passage ranker as the one in MUSST.

\subsection{Implementation details}
For the multi-span answer annotation, we use constituency parser from Standford CoreNLP \cite{manning_stanford_2014}. NLTK \footnote{\url{https://www.nltk.org}} package is also used to implement our annotator. The maximum edit distance between the answer reconstructed from the annotated spans, and the original answer is 32 and 8 respectively for the $\mathcal{NLG}$ and $\mathcal{QA}$ training sets.

The ranker and question-answering module of MUSST are implemented with PyTorch \footnote{\url{https://pytorch.org}} and Transformers package \footnote{\url{https://github.com/huggingface/transformers}}. We adopt ALBERT \cite{lan_albert_2020} as the encoder in our models and initialize it with the pre-trained weights before the fine-tuning. We choose ALBERT-base as the encoder of passage ranker and ALBERT-xlarge instead for question answering module.

Following \citeauthor{lan_albert_2020}~(\citeyear{lan_albert_2020}), we use SentencePiece \cite{kudo_sentencepiece_2018} to tokenize our inputs with a vocabulary size of 30,000. We adopt Adam optimizer \cite{kingma_adam_2015} to minimize the cost function. Two types of regularization methods during training: dropout and L2 weight decay. Hyperparameter details for the training of the different models of our framework are presented in Table \ref{tab:hyperparameters}. MUSST-NLG and MUSST-QA are trained respectively on the $\mathcal{NLG}$ and $\mathcal{QA}$ subsets. The maximum number of spans for them is set to 9 and 5, respectively. We trained the passage ranker and the question-answering module of MUSST-NLG on a machine with four Tesla P40 GPUs. The question-answering module of MUSST-QA is trained with eight GeForce GTX 1080 Ti GPUs. It takes roughly 9 hours to train the passage ranker. For the question-answering module in MUSST-NLG and MUSST-QA, the training time is about 10 hours and 17 hours respectively.
\begin{table}[!htbp]
	\centering
	\resizebox{0.48\textwidth}{!}{
		\begin{tabular}{lccccccccc}
			\toprule
			\textbf{Hyperparameter} & \textbf{Ranker} & \textbf{MUSST-QA} & \textbf{MUSST-NLG} \\
			\midrule
			Learning rate           & 1e-5            & 3e-5              & 3e-5               \\
			Learning rate decay     & Linear          & Linear            & Linear             \\
			Training epoch          & 3               & 3                 & 5                  \\
			Warmup rate             & 0.1             & 0.1               & 0.1                \\
			Adam $\epsilon$         & $10^{-6}$       & $10^{-6}$         & $10^{-6}$          \\
			Adam $\beta_1$          & 0.9             & 0.9               & 0.9                \\
			Adam $\beta_2$          & 0.999           & 0.999             & 0.999              \\
			MSN                     & 256             & 256               & 256                \\
			Batch size              & 128             & 32                & 32                 \\
			Encoder dropout rate    & 0               & 0                 & 0                  \\
			Classifier dropout rate & 0.1             & 0.1               & 0.1                \\
			Weight decay            & 0.01            & 0.01              & 0.01               \\
			\bottomrule
		\end{tabular}
	}
	\caption{Training hyperparameters of different modules of MUSST on MS MARCO v2.1 dataset.  Here, MUSST-QA and MUSST-NLG refer to its question-answering module. MSN means maximum sequence length.}
	\label{tab:hyperparameters}
\end{table}

The single-span baseline is implemented with the same packages as MUSST while the seq2seq baseline is implemented with Fairseq \cite{ott2019fairseq}.

\subsection{Results}
\begin{table}[!htbp]
	\centering
	\resizebox{0.47\textwidth}{!}{
		\begin{tabular}{lcccc}
			\toprule
			\multirow{2}{*}{\textbf{Model} } & \multicolumn{2}{c}{$\mathcal{QA}$} & \multicolumn{2}{c}{$\mathcal{NLG}$}                                      \\
			                                 & \textbf{ROUGE-L}                   & \textbf{BLEU-1}                     & \textbf{ROUGE-L} & \textbf{BLEU-1} \\
			\midrule
			% \multicolumn{5}{c}{\textit{Baseline}} \\
			Single-span                      & 47.96                              & \textbf{50.22}                      & 53.10            & 49.08           \\
			Seq2seq                          & --                                 & --                                  & 56.42            & 53.89           \\
			\midrule
			MUSST-QA                         & \textbf{48.44}                     & 49.54                               & --               & --              \\
			MUSST-NLG                        & --                                 & --                                  & \textbf{66.24}   & \textbf{64.23}  \\
			\bottomrule
		\end{tabular}
	}
	\caption{Performance comparison with our baselines on the $\mathcal{QA}$ and $\mathcal{NLG}$ development set. Here, we use the same single ranker for MUSST and the baselines.}
	\label{tab:dev_baseline}
\end{table}

\begin{table*}[!htbp]
	\centering
	\resizebox{\textwidth}{!}{
		\begin{tabular}{lccccccc}
			\toprule
			\multirow{2}{*}{\textbf{Model}} & \multirow{2}{*}{\textbf{Answer Generation}} & \multirow{2}{*}{\textbf{Ranking}} & \multicolumn{2}{c}{\textbf{NLG Task}} & \multicolumn{2}{c}{\textbf{QA Task}} & \multirow{2}{*}{\textbf{Overall Average}}                                   \\
			                                &                                             &                                   & \textbf{R-L}                          & \textbf{B-1}                         & \textbf{R-L}                              & \textbf{B-1}   &                \\
			\midrule
			Human                           & --                                          & --                                & 63.2                                  & 53.0                                 & 53.9                                      & 48.5           & 54.65          \\
			\hdashline
			\multicolumn{8}{c}{\textit{Unpublished}}                                                                                                                                                                                                                                       \\
			PALM                            & \multicolumn{2}{c}{Unknown}                 & \textbf{49.8}                     & 49.9                                  & 51.8                                 & 50.7                                      & \textbf{50.55}                  \\
			Multi-doc Enriched BERT         & \multicolumn{2}{c}{Unknown}                 & 32.5                              & 37.7                                  & \textbf{54.0}                        & \textbf{56.5}                             & 45.18                           \\
			\hdashline
			\multicolumn{8}{c}{\textit{Published}}                                                                                                                                                                                                                                         \\
			BiDAF$^a$ \spade                & Single-span                                 & Confidence score                  & 16.9                                  & 9.3                                  & 24.0                                      & 10.6           & 15.20          \\
			ConZNet$^b$  \spade             & Pointer-Generator                           & Unkonwn                           & 42.1                                  & 38.6                                 & --                                        & --             & --             \\
			VNET$^c$  \spade                & Single-span                                 & Answer verification               & 48.4                                  & 46.8                                 & 51.6                                      & 54.3           & \textbf{50.28} \\
			Deep Cascade QA$^d$  \spade     & Single-span                                 & Cascade                           & 35.1                                  & 37.4                                 & 52.0                                      & 54.6           & 44.78          \\
			Masque QA$^e$ \td               & Pointer-Generator                           & Joint trained classifier          & 28.5                                  & 39.9                                 & 52.2                                      & 43.7           & 41.08          \\
			Masque NLG$^e$ \td              & Pointer-Generator                           & Joint trained classifier          & 49.6                                  & \textbf{50.1}                        & 48.9                                      & 48.8           & 49.35          \\
			\midrule
			\textbf{MUSST-NLG} \td          & Multi-span                                  & Standalone classifier             & 48.0                                  & 45.8                                 & 49.0                                      & 51.6           & 48.60          \\
			\bottomrule
		\end{tabular}}
	\caption{The performance of our framework and competing models on the MS MARCO v2.1 test set. All the results presented here reflect the MS MARCO leaderboard (\url{microsoft.github.io/msmarco/}) as of 28 May 2020. $\spadesuit$ refers to the model whose results are not reported in the original published paper. BiDAF for MARCO is implemented by the official MS MARCO Team. \textdagger \, refers to the ensemble submission. Whether the other competing models are ensemble or not is unclear.
		$^a$ \citeauthor{seo_bidirectional_2017}~(\citeyear{seo_bidirectional_2017});
		$^b$ \citeauthor{indurthi_cut_2018}~(\citeyear{indurthi_cut_2018});
		$^c$ \citeauthor{wang_multi-passage_2018}~(\citeyear{wang_multi-passage_2018});
		$^d$ \citeauthor{yan_deep_2019}~(\citeyear{yan_deep_2019});
		$^e$ \citeauthor{nishida_multi-style_2019}~(\citeyear{nishida_multi-style_2019}).
	}
	\label{tab:test}
\end{table*}

Table \ref{tab:dev_baseline} shows the results of our single model and the baseline models on the $\mathcal{QA}$ and $\mathcal{NLG}$ development datasets. MUSST outperforms significantly the baselines including the generative seq2seq model over the $\mathcal{NLG}$ set in terms of both ROUGE-L and BLEU-1. Even on the $\mathcal{QA}$ set, our model yields better results regarding ROUGE-L. Table \ref{tab:test} compares our model performance with the competing models on the leaderboard. Although our model utilizes only a standalone classifier for passage ranking, multi-span style extraction still helps us rival with state-of-the-art approaches.

\section{Analysis and discussions}
\subsection{Effect of maximum number of spans}
\begin{figure}[!htbp]
	\centering
	\includegraphics[width=\linewidth]{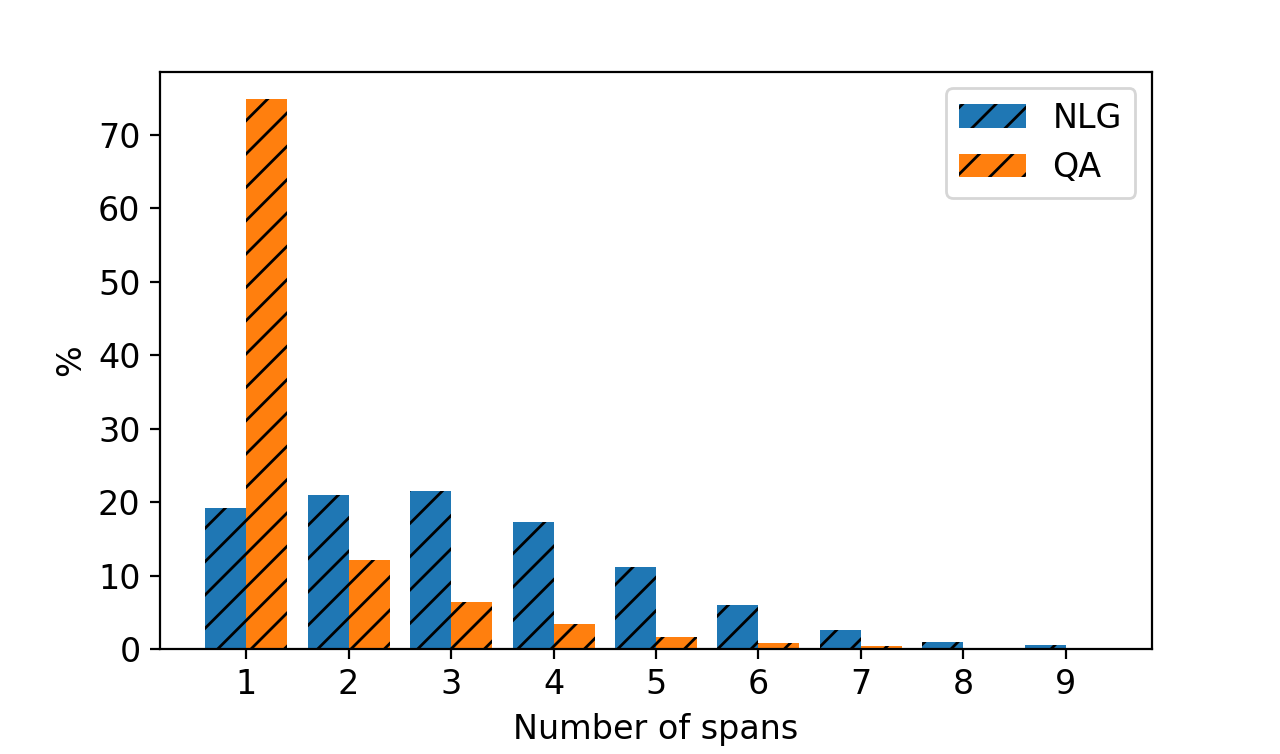}
	\caption{Distribution of training samples of edit distance less than 4 over annoted answer spans. For the purpose of better illustration, we filter the samples which include more than 9 spans.}
	\label{fig:span_dist}
\end{figure}

Figure \ref{fig:span_dist} presents the distribution of span numbers with edit distance less than 4 over the $\mathcal{QA}$ and $\mathcal{NLG}$ training sets after the annotation procedure. It is seen that most QA-style answers are only one span, while the NLG-style answers distribute more uniformly in the range of [1, 9].

\begin{figure}[!htbp]
	\centering
	\includegraphics[width=\linewidth]{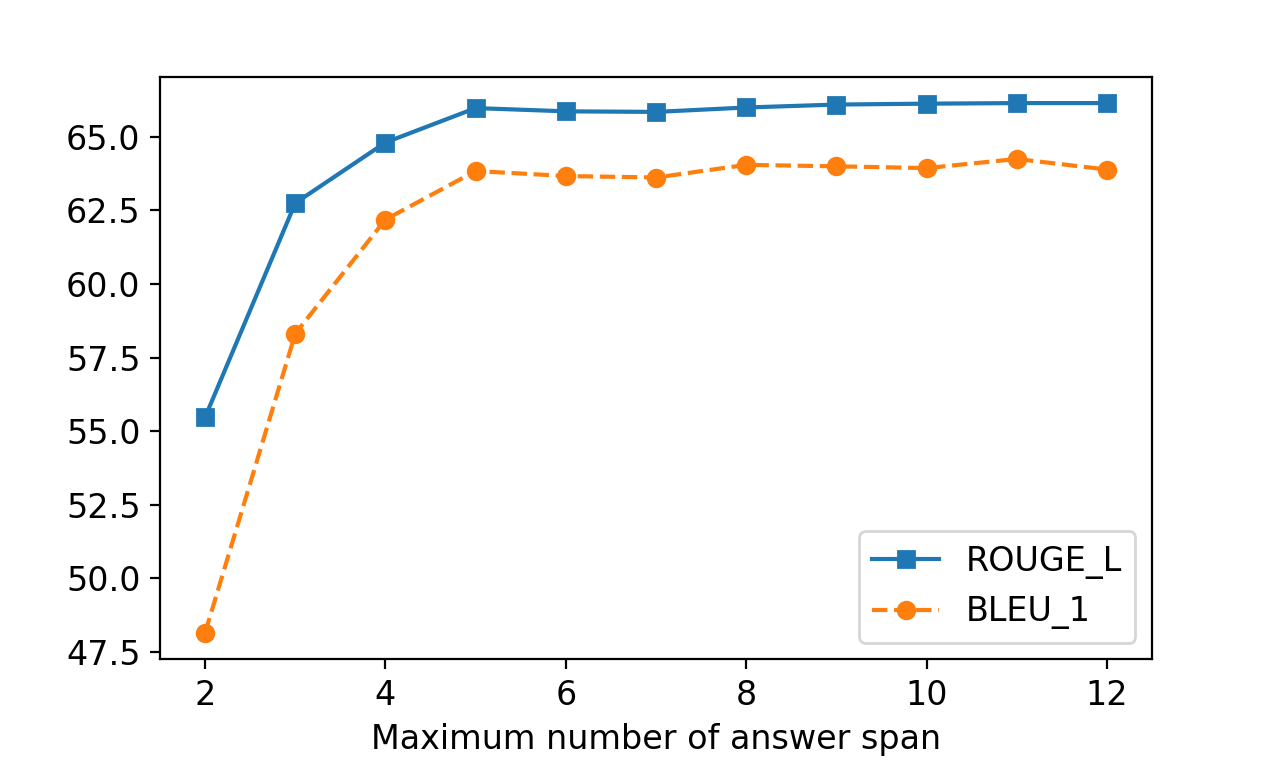}
	\caption{Effect of maximum number of spans.}
	\label{fig:effet_mns}
\end{figure}

To better understand the effect of the maximum number of spans to be generated in the answer generator, we let it vary in the range of [2, 12] and conduct experiments on the $\mathcal{NLG}$ set with our best single passage ranker. The edit distance threshold is set to be 8. The results are presented in Figure \ref{fig:effet_mns}. Generally, increasing the number of the span will augment the token coverage rate, thus yielding better results. But the gain becomes less significant when the maximum number of span is already large enough. From Figure \ref{fig:effet_mns}, we can see that the results vary imperceptibly when the maximum number of spans reaches 5. However, since each span only introduces 4k parameters, which is negligible before the encoder (60M), we still choose the maximum number to be 9, which corresponds to the best performance on the development set.

\subsection{Ablation study on model design choice}
We perform ablation experiments that quantify the individual contribution of the design choices of MUSST. Table \ref{tab:ablation} shows the results on the $\mathcal{NLG}$ development set. Both \textit{pruning} and \textit{conditional masking} contribute the model performance, which indicates that pruning can help the model to converge more easily by reducing the number of spans, while conditional masking can better generate answer without suffering from the repeating problem. We also observe using the gold passage can significantly improve question-answering. It shows there still exists a great improvement space for the passager ranker.
\begin{table}[!htbp]
	\centering
	% \resizebox{0.45\textwidth}{!}{
	\begin{tabular}{lcccc}
		\toprule
		\textbf{Model}
		                           & \textbf{ROUGE-L} & \textbf{BLEU-1} \\
		\midrule
		MUSST                      & 66.24            & 64.23           \\
		\, w/o pruning             & 64.66            & 60.36           \\
		\, w/o conditional masking & 65.50            & 64.31           \\
		\midrule
		MUSST w gold passage       & 75.39            & 74.41           \\
		\bottomrule
	\end{tabular}
	% }
	\caption{Ablation study on the $\mathcal{NLG}$ development set.}
	\label{tab:ablation}
\end{table}

\subsection{Quality of multi-span answer annotator}
On the $\mathcal{NLG}$ development set, we evaluate the answers generated by our syntactic multi-span annotator. The results shows our annotated answers can obtain 89.35 in BLEU-1 and 90.19 in ROUGE-L with the gold passages, which demonstrates the effectiveness of our annotator. For MUSST, the results are 74.41 and 75.39 respectively (in Table \ref{tab:ablation}). So there is still much room for improvement with respect to the question-answering module.

\subsection{Effect of edit distance threshold}
\begin{figure}[!htbp]
	\centering
	\includegraphics[width=1\linewidth]{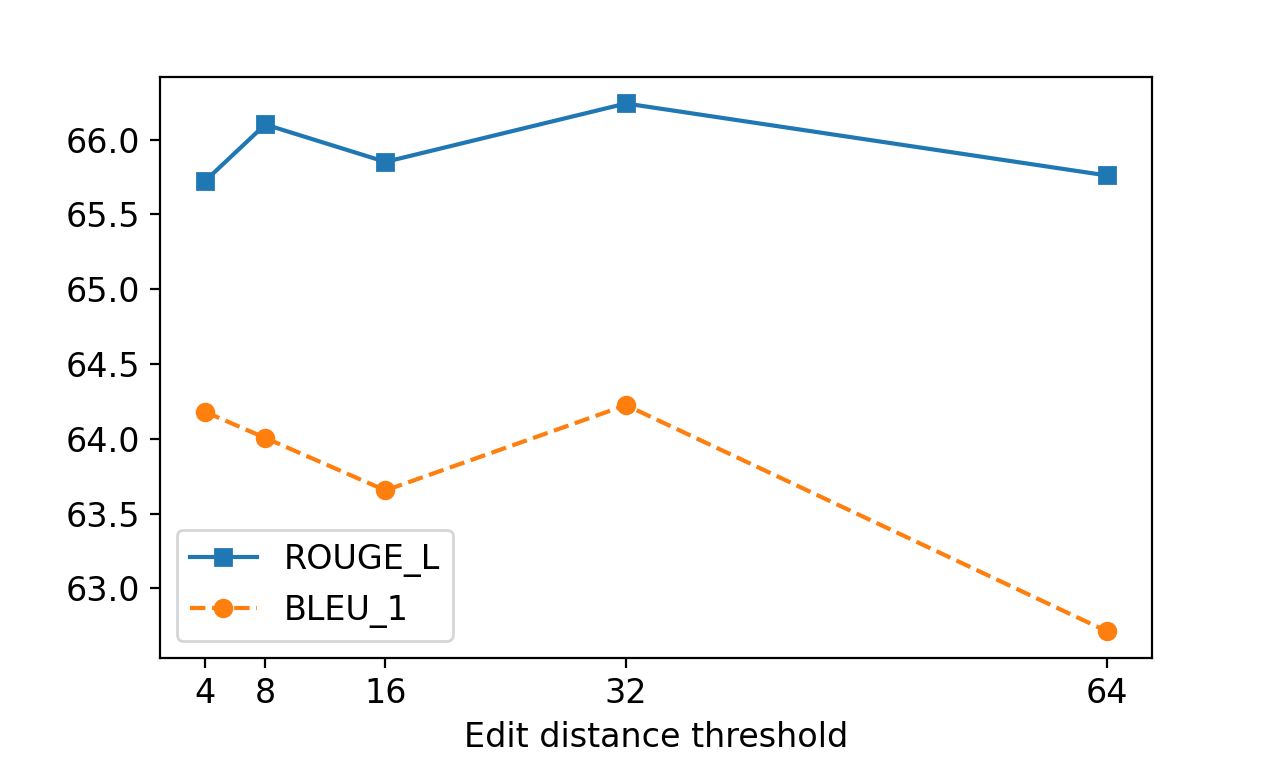}
	\caption{Effect of edit distance threshold.}
	\label{fig:effet_ed}
\end{figure}

Figure \ref{fig:effet_ed} shows the results of MUSST on $\mathcal{NLG}$ development set for various edit distance threshold. Interestingly, it indicates that BLEU-1 is impacted more heavily by the variation of edit distance than ROUGE-L. And setting the edit distance threshold too large may damage the model performance by introducing too many \textit{incomplete} samples.

\subsection{Effect of encoder size}
Table \ref{tab:encoder} presents experimental results on ALBERT encoder with various model sizes. Unsurprisingly, the model yields stronger results as the encoder gets larger.

\begin{table}[!htbp]
	\centering
	\resizebox{0.45\textwidth}{!}{
		\begin{tabular}{lcccc}
			\toprule
			\textbf{Encoder}
			              & \textbf{Parameters} & \textbf{ROUGE-L} & \textbf{BLEU-1} \\
			\midrule
			ALBERT-base   & 12M                 & 62.03            & 60.48           \\
			ALBERT-large  & 18M                 & 64.93            & 61.67           \\
			ALBERT-xlarge & 60M                 & \textbf{66.24}   & \textbf{64.23}  \\
			\bottomrule
		\end{tabular}
	}
	\caption{Effect of ALBERT encoder size.}
	\label{tab:encoder}
\end{table}

\subsection{Performance of the ranker}
\begin{table}[!htbp]
	\centering
	\resizebox{0.48\textwidth}{!}{
		\begin{tabular}{lccc}
			\toprule
			\textbf{Model}          & \textbf{Training set} & \textbf{MAP}   & \textbf{MRR}   \\
			\midrule
			Bing (initial ranking)  & -                     & 34.62          & 35.00          \\
			\midrule
			\textbf{MUSST} (single) & $\mathcal{QA}$        & \textbf{71.10} & \textbf{71.56} \\
			\, w/o dynamic sampling & $\mathcal{QA}$        & 70.82          & 71.26          \\
			\bottomrule
		\end{tabular}
	}
	\caption{The performance of ranker with various configurations on the $\mathcal{QA}$ development set.}
	\label{tab:ranker}
\end{table}
Table \ref{tab:ranker} presents our ranker performance in terms of MAP and MRR. The results show that dynamic sampling leads to slightly better results.

\subsection{Case study}

To have an intuitive observation of the prediction ability of MUSST, we show a prediction example on MS MARCO v2.1 from the baseline and MUSST in Table \ref{tab:case}. The comparison indicates that our model could extract effectively useful spans, yielding more complete answer that can be understood independent of question and passage context.
\begin{table}[htbp!]
	\centering
	\begin{tabular}{l}
		\toprule
		\pbox{0.45\textwidth}{
		\textbf{Question}: \textit{how long should \hl{a central air conditioner} last}                                                                                                                                                                                                                                                                                                                                             \\
		\textbf{Selected Passage}: \textit{\hl{10 to 20 years} - sometimes longer. You \hl{should} have a service tech come out once a year for a tune up. You wouldn't run your car without regular maintenance and tune ups and you shouldn't run your a/c that way either - if you want it to \hl{last} as long as possible. Source(s): 20 years working \hl{for} a major manufacturer of central heating and air conditioning.} \\
		\textbf{Reference Answer}: \textit{A Central air conditioner lasts for in between 10 and 20 years./ A central air conditioner should last for 10 to 20 years. }                                                                                                                                                                                                                                                             \\
		\textbf{Prediction (Baseline)}: \textit{10 to 20 years.}                                                                                                                                                                                                                                                                                                                                                                    \\
			\textbf{Prediction (MUSST)}: \textit{a central air conditioner should last for 10 to 20 years.}
		}                                                                                                                                                                                                                                                                                                                                                                                                                           \\
		\bottomrule
	\end{tabular}
	\caption{A prediction example from the baseline and MUSST. The underlined texts are the spans predicted by our model to compose the final answer phrase.}
	\label{tab:case}
\end{table}

\section{Related work}
\subsection{Generative MRC}
Generative MRC is considered as a more challenging task where answers are free-form human-generated text. More recently, we have seen an emerging wave of generative MRC tasks, including MS MARCO \cite{bajaj_ms_2018}, NarrativeQA \cite{kocisky_narrativeqa_2018}, DuReader \cite{he_dureader_2018} and CoQA \cite{reddy_coqa_2019}.

The most earlier approaches tried to generate the answer in a single-span extractive way \cite{tay_densely_2018, tay_multi-granular_2018, wang_multi-passage_2018, yan_deep_2019, ohsugi_simple_2019}. The models using a single-span extractive method show effectiveness for the dataset where abstractive behavior of answers includes mostly small modifications to spans in the context \cite{ohsugi_simple_2019, yatskar_qualitative_2019}. Whereas, for the datasets with answers of deep abstraction, this method fails to yield promising results. The first attempt to generate the answer in a generative way is to apply an RNN-based seq2seq attentional model to synthesize the answer, such as S-NET \cite{tan_s-net_2018}, where seq2seq learning was first introduced by \citeauthor{sutskever_sequence_2014}~(\citeyear{sutskever_sequence_2014}) for the machine translation. The most recent models adopt a hybrid neural network Pointer-Generator \cite{see_get_2017} to generate answer, such as ConZNet \cite{indurthi_cut_2018}, MHPGM \cite{bauer_commonsense_2018} and Masque \cite{nishida_multi-style_2019}. Pointer-Generator was firtsly proposed for the abstractive text summarization, which can copy words from the source via the pointer network while retaining the ability to produce novel words through the generator. Different from ConZNet and MHPGM, Masque adopt a Transformer-based \cite{vaswani_attention_2017} Pointer-Generator, while the previeous ones utilizeing GRU \cite{cho_learning_2014} or LSTM \cite{hochreiter_long_1997}.

\subsection{Multi-passage MRC}
For each question-answer pair, the Multi-passage MRC dataset contains more than one passage as the reading context, such as SearchQA \cite{dunn_searchqa_2017}, Triviaqa \cite{joshi_triviaqa_2017}, MS MARCO, and DuReader.

Existing approaches designed specifically for Multi-passage MRC can be classified into two categories: pipeline and end-to-end. Pipeline-based models \cite{chen_reading_2017, wang_r3_2018, clark_simple_2018} adopt a ranker to first rank all the passages based on its relevance to the question and then utilize a question-answering module to read the selected passages. The ranker can be based on traditional information retrieval methods (BM25 or TF-IDF) or employ a neural re-ranking model. End-to-end models \cite{wang_multi-passage_2018, tan_s-net_2018, nishida_multi-style_2019} read all the provided passages at the same time, and produce for each passage a candidate answer assigned with a score which is consequently compared among passages to find the final answer. Passage ranking and answer prediction are usually jointly done as multi-task learning. More recently, \citeauthor{yan_deep_2019}~(\citeyear{yan_deep_2019}) proposed a cascade learning model to balance the effectiveness and efficiency of the two approaches mentioned above.

\subsection{Pre-trained language model in MRC}
Employing the pre-trained language models has been a common practice for tackling MRC tasks \cite{zhang_machine_2020}. The appearances of more elaborated architectures, larger corpora, and more well-designed pre-training objectives speed up the achievement of new state-of-the-art in MRC \cite{devlin_bert:_2019, liu_roberta:_2019, yang_xlnet:_2019, lan_albert_2020}. Moreover, \citeauthor{glass_span_2019}~(\citeyear{glass_span_2019}) adopts span selection, a MRC task, as an auxiliary pre-training task. Another mainstream line of research attempts to drive the improvements during the fine-tuning, which includes integrating better verification strategies for unanswerable question \cite{zhang_retrospective_2020}, incorporating explicit linguistic features \cite{zhang_semantics-aware_2020, zhang_sg-net_2020}, leveraging external knowledge for commonsense reasoning \cite{lin_kagnet_2019} or enhancing matching network for multi-choice MRC \cite{zhang_dcmn_2020, zhu_dual_2020}. In addition,\citeauthor{hu_multi-type_2019}~(\citeyear{hu_multi-type_2019}) introduced multi-span extraction to obtain top-k most likely spans for multi-type MRC. However, different from our work, this method is more suitable to predict a set of independent answer spans instead of generating a complete sentence.

\section{Conclusion}
In this work, we present a novel solution to generative MRC, multi-span style extraction framework (MUSST), and show it is capable of alleviating the problems of generating incomplete answers or introducing redundant words encountered by single-span extraction models. We apply our model to a challenging generative MRC dataset MS MARCO v2.1 and significantly outperform the single-span extraction baseline. This work indicates a new research line for generative MRC in addition to the existing two methods, single-span extraction and seq2seq generation. With the support of only a standalone ranking classifier, our proposed method still gives an overall performance approaching state-of-the-art, showing great potential.

\bibliography{musst}

\end{document}

%% file: math_commands.tex
\usepackage{amsmath,amsfonts,bm}

\def\eqref#1{equation~\ref{#1}}
\def\1{\bm{1}}

\def\vtheta{{\bm{\theta}}}

\def\vb{{\bm{b}}}
\def\vc{{\bm{c}}}

\def\vp{{\bm{p}}}

\def\vs{{\bm{s}}}

\def\mE{{\bm{E}}}

\def\mM{{\bm{M}}}

\def\mW{{\bm{W}}}

\DeclareMathAlphabet{\mathsfit}{\encodingdefault}{\sfdefault}{m}{sl}
\SetMathAlphabet{\mathsfit}{bold}{\encodingdefault}{\sfdefault}{bx}{n}

\def\gR{{\mathcal{R}}}

\def\gT{{\mathcal{T}}}
\def\gU{{\mathcal{U}}}
\def\gV{{\mathcal{V}}}

\def\sS{{\mathbb{S}}}